%% file: main.tex
\documentclass[letterpaper, 10 pt, conference]{ieeeconf}  

\IEEEoverridecommandlockouts                              

\usepackage{graphics} 
\usepackage{epsfig} 
\usepackage{mathptmx} 
\usepackage{times} 
\usepackage{amsmath} 
\usepackage{amssymb}  
\usepackage{booktabs}
\usepackage{color}
\usepackage{hyperref}
\hypersetup{
    colorlinks=true,
    linkcolor=blue,
    urlcolor=blue,
    citecolor=blue
}
\usepackage{xcolor}
\usepackage[font=small]{caption}

\begin{document}

\definecolor{pine}{RGB}{62,145,103}
\definecolor{myblue}{RGB}{50,50,255}

\newcommand{\BJ}[1]{\textcolor{pine}{[BJ: #1]}} 
\newcommand{\jmf}[1]{\textcolor{myblue}{[JMF: #1]}}
\newcommand{\NK}[1]{\textcolor{purple}{[NK: #1]}}

\title{ 
ProbeMDE: Uncertainty-Guided Active Proprioception for \\ Monocular Depth Estimation in Surgical Robotics
}
\author{ 
    Britton Jordan*$^{1}$,
    Jordan Thompson*$^{1}$,
    Jesse F. d'Almeida$^{2}$,
    Hao Li$^{3}$,
    Nithesh Kumar$^{2}$,
    Susheela Sharma Stern$^{2}$,\\
    James Ferguson$^{1}$,
    Ipek Oguz$^{3}$,
    Robert J. Webster III$^{2}$,
    Daniel Brown$^{1}$, and
    Alan Kuntz$^{3, 4}$
\thanks{*These authors contributed equally.}
\thanks{$^{1}$The Robotics Center and the Kahlert School of Computing at the University of Utah, Salt Lake City, UT 84112, USA.}
\thanks{The Departments of $^{2}$Mechanical Engineering, $^{3}$Computer Science, and $^{4}$Electrical and Computer Engineering at Vanderbilt University, Nashville, TN, 37203, USA.}
\thanks{Research reported in this publication was supported by the Advanced Research Projects Agency for Health (ARPA-H) under Award Number D24AC00415-00 for ALISS. The ARPA-H award provided 90\% of total costs with an award total of up to \$10,741,534.20 and the National Science Foundation (NSF) Graduate Research Fellowship Program funds 10\% of total costs under Grant No. 2139322. The content is solely the responsibility of the authors and does not necessarily represent the official views of \mbox{ARPA-H} or the NSF.}
}

\maketitle

\thispagestyle{empty}
\pagestyle{empty}

\input{0_Abstract}
\input{1_Introduction}
\input{2_RelatedWork}
\input{3_ProblemDefinition}
\input{4_Methodology}
\input{5_Experiments}
\input{6_Results}

\section{Limitations and Future Work}
While our approach demonstrates clear improvements in monocular depth estimation through uncertainty-guided active sensing, several limitations remain. For instance, our method assumes a static environment, whereas the endoscopic scene will change as the surgery is conducted. In the future, we plan to account for temporal dynamics by considering the historical sequence of depth measurements and endoscopic images. Another limitation of relying heavily on robot probes is the issue of kinematic reachability. Since not every point in the endoscopic image is reachable by the robot, we are unable to obtain depth measurements in some areas. Future work could explore tracked repositioning of the surgical robot base to enable probing areas which were previously unreachable. In this work we demonstrate that a U-Net trained for a targeted endoscopic procedure undergoes major performance gains with the fusion of sparse proprioceptive measurements. Future work in MDE could benefit from using our proposed framework to incorporate sparse depth measurements into larger foundation models for improved absolute depth accuracy both in surgery and across many domains.


\input{7_Conclusion}


\bibliographystyle{ieeetr}
\bibliography{citations}

\end{document}

%% file: 0_Abstract.tex
\begin{abstract}

Monocular depth estimation (MDE) provides a useful tool for robotic perception, but its predictions are often uncertain and inaccurate in challenging environments such as surgical scenes where textureless surfaces, specular reflections, and occlusions are common. To address this, we propose ProbeMDE, a cost-aware active sensing framework that combines RGB images with sparse proprioceptive measurements for MDE. Our approach utilizes an ensemble of MDE models to predict dense depth maps conditioned on both RGB images and a sparse set of known depth measurements obtained via proprioception, where the robot has touched the environment in a known configuration. We quantify predictive uncertainty via the ensemble's variance and measure the gradient of the uncertainty with respect to candidate measurement locations. To prevent mode collapse while selecting maximally informative locations to propriocept (touch), we leverage Stein Variational Gradient Descent (SVGD) over this gradient map. We validate our method in both simulated and physical experiments on central airway obstruction surgical phantoms. Our results demonstrate that our approach outperforms baseline methods across standard depth estimation metrics, achieving higher accuracy while minimizing the number of required proprioceptive measurements. The project website is \mbox{\href{https://brittonjordan.github.io/probe_mde/}{brittonjordan.github.io/probe\_mde}}
\end{abstract}

%% file: 1_Introduction.tex
\section{Introduction}

Depth perception is a fundamental requirement for robotic systems operating in unstructured and deformable environments such as surgical settings. Monocular depth estimation (MDE) offers one potential solution due to its ability to infer dense 3D scene structure from a single RGB camera without the need for specialized depth sensors~\cite{ming2021deep}. However, MDE models often struggle with ambiguities arising from textureless surfaces, specular reflections, and occlusions, all of which are especially prevalent in robotic surgical settings~\cite{daher2025shades}.

One avenue to address these limitations is to combine vision-based predictions with proprioceptive measurements~\cite{ferguson2020toward},~\cite{Kumar_Hamlyn2025}. We define robotic proprioception as a robot's ability to sense contact with external objects in addition to an awareness of its own position in space through tool-tip kinematics. Utilizing proprioception, we can provide sparse ground-truth depth measurements at contact points between the robot and the environment. While these measurements can provide accurate estimates of depth at specified locations, their acquisition has a cost as each proprioceptive measurement requires physical interaction with the environment that will slow down task execution. Thus, we seek to minimize the number of required measurements while maximizing the reconstruction accuracy of the scene.

To solve this problem, we introduce ProbeMDE, a cost-aware active sensing framework for monocular depth estimation. Our approach leverages an ensemble of U-Net-based MDE models trained to predict dense depth maps from RGB images of a surgical scene conditioned on a sparse set of propriocepted depth measurements. We quantify predictive uncertainty by computing the variance across ensemble predictions to produce a differentiable scalar uncertainty measure. By analyzing the gradient of this measure with respect to potential depth measurement locations, we then employ Stein Variational Gradient Descent (SVGD) to optimize the selection of query points while minimizing mode collapse.

\begin{figure}[!t]
    \centering
    \includegraphics[width=1.0\linewidth]{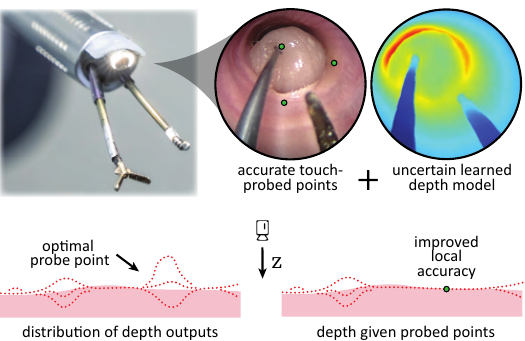}
    \caption{Conceptual illustration of our approach to fuse probabilistic learned depth models with touch-based position data. With no probed points (bottom, left), depth estimation can be uncertain. Our approach chooses the acquisition point that will minimize variance, and then re-estimates iteratively.}
    \vspace{-1.5em}
    \label{fig:front_page}
\end{figure}

We validate our framework in both simulated and physical experiments. In simulation, where full ground-truth depth maps are available, we benchmark our approach against several baseline point selection methods including random sampling, greedy heuristics, and gradient-based heuristics. Our method consistently achieves greater reconstructive performance across standard depth estimation metrics. In physical experiments inside surgical phantoms, we demonstrate our method's reconstructive accuracy and robustness over baseline approaches.

Our main contributions are twofold. First, we present a cost-aware active sensing framework that conditions monocular depth estimation on sparse proprioceptive depth measurements, including the first use of proprioception to improve monocular depth estimation in robotic endoscopic surgery. Second, we introduce a principled and efficient strategy for selecting depth measurement locations that leverages SVGD applied to the gradient of ensemble predictive variance, enabling targeted sensing that improves depth estimates and efficient measurement acquisition.



\begin{figure*}
    \centering
    \includegraphics[width=0.95\textwidth]{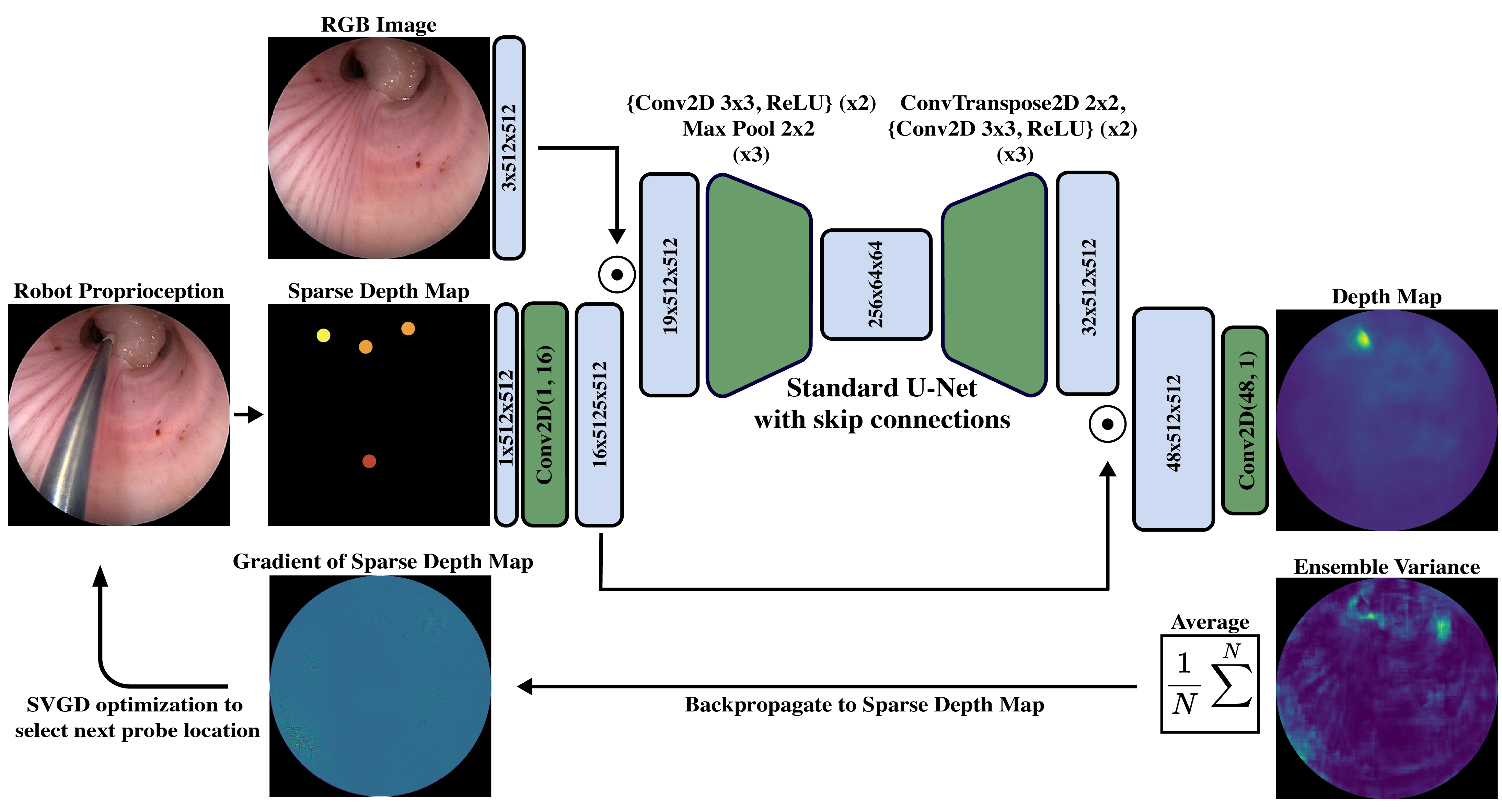}
    \caption{ProbeMDE, a cost-aware active sensing framework for monocular depth estimation. Inputs are an endoscopic RGB image and a sparse propriocepted depth map $S$. We upsample $S$ to 16 channels with a 2D convolution before concatenating with the image and passing through a U-Net encoder. Using 3 successive iterations of 2D convolutions, ReLU activation, and max pooling, the input is transformed to a 64x64 matrix with 256 channels. The convolution transpose is used to expand back to 512x512 while a 2D convolution and ReLU activation are used to combine skip connections. The upsampled sparse ground truth depth is concatenated to the output of the U-Net's last layer before a final convolution, giving the network additional opportunity to leverage this ground-truth information. The network produces a dense 512x512 depth map. We compute the variance map $U$ of the ensemble's output, from which we compute $\nabla_{S}\lVert U\rVert$. Locations to probe next are selected using SVGD on the gradient map and then propriocepted with the surgical robot. The new depths are added to the sparse ground truth and the process is re-run.}

    \label{fig:system_architecture}
    \vspace{-1.5em}
\end{figure*}

%% file: 2_RelatedWork.tex
\section{Related Work}

\subsection{Monocular Depth Estimation}

Monocular depth estimation uses an image from a single camera viewpoint to predict the distance from the camera to each point in the 3D scene. State-of-the-art MDE methods produce depth maps with outstanding accuracy for a wide variety of everyday indoor and outdoor scenes \cite{yang_depth_2024}, \cite{ke_repurposing_2024}. However, endoscopic videos of surgical scenes are out of distribution from the large-scale datasets these foundation models are trained on, resulting in suboptimal performance~\cite{shao_endomust_2025}.

Efforts to improve MDE in endoscopy include adapting or fine-tuning foundation models with simulation data \cite{shao_endomust_2025}, \cite{tan_accurate_2025}, \cite{li_monocular_2025}, using a style transfer network to make simulation renderings more photorealistic \cite{jeong_depth_2024}, or utilizing point cloud data from structure from motion as sparse-ground-truth depth cues \cite{liu_dense_2020}. MDE models focused on a specific domain frequently use an adaptation of the U-Net architecture \cite{ronneberger2015u}, \cite{jeong_depth_2024}. In this work, we draw from a similar toolbox of approaches while adding the ability to adaptively improve depth prediction through uncertainty-guided active sensing. As in prior works, we utilize a style-transfer network to make simulation RGB renderings more realistic. We then use this data to fine-tune a large foundation model, Depth Anything V2 \cite{yang_depth_2024}, which we subsequently use to generate pseudo-ground-truth data for training a model with better real-world performance. We introduce a novel source of cues for endoscopic MDE, robotic proprioception at specific locations within the scene which will maximally resolve the uncertainty of our prediction.

\subsection{Uncertainty Quantification}

As deep neural networks become more complex and less interpretable, uncertainty quantification has become a critical component of any such system, especially for safety-critical tasks such as robotic surgery. Many methods exist for quantifying predictive uncertainty in deep learning including deep ensembles \cite{lee2015m}, Monte Carlo dropout \cite{milanes2021monte}, \cite{camarasa2020quantitative}, \cite{mae2021uncertainty}, and test-time augmentation \cite{wang_aleatoric_2019}. A comparison of these methods shows that deep ensembles consistently provide the most accurate quantification of uncertainty under distribution shift in the data~\cite{ovadia_can_2019}. An ensemble is created by training multiple models with different random weight initializations such that they each converge to a different local minima in the loss space. At inference time, we can use the variance of the ensemble members' predictions as a proxy for uncertainty ~\cite{fort2019deep},~\cite{pearce2018high},~\cite{lakshminarayanan2017simple}. Many works have quantified the uncertainty of MDE predictions~\cite{xiang_measuring_2024},~\cite{marsal_monoprob_2024},~\cite{landgraf_critical_2025}, including some focused specifically on endoscopic images~\cite{chong_monocular_2025}. However, no past works have explored the active refinement of MDE predictions using robotic touch sensing. In this work, we leverage uncertainty quantification techniques for ensembles to develop an uncertainty-guided active touch sensing framework for monocular depth estimation.

\subsection{Active Touch Sensing}

Active touch sensing has been widely explored as a means of acquiring task-relevant information under sensing constraints. Early approaches focused on reconstructing object geometry through sparse and targeted contact interactions For instance, tactile fingertips have been used to actively trace object contours for shape recognition~\cite{assaf2014seeing}, while force-based probing has enabled reconstruction of unknown objects~\cite{kim2015exploration}. More recent work has emphasized uncertainty-driven exploration, where sensing actions are guided to reduce ambiguity~\cite{martinez2017active}. Recent research has further advanced active touch sensing with machine learning and task-specific exploration. Learning-based approaches have been proposed to determine not only how to explore unknown objects, but also learning where to touch and how to recognize objects from tactile data~\cite{shahidzadeh2024actexplore,xu2022tandem}. Together, these works demonstrate the value of active touch sensing to improve perception and task performance. In contrast, our work extends the concept of active touch sensing into the domain of proprioceptive depth acquisition, where sparse depth measurements are strategically chosen to reduce uncertainty in monocular depth estimation for deformable surgical environments.

%% file: 3_ProblemDefinition.tex
\section{Problem Definition}

We formalize the task of selecting a sparse set of proprioceptive measurements as a cost-constrained uncertainty minimization problem for monocular depth estimation. Let $I\in\mathbb{R}^{HxWx3}$ be the RGB image of a surgical scene, and let $\mathrm{S}=\{(x_i,y_i,d_i)\}_{i=1}^N$ be a sparse set of ground-truth depth measurements where each location $(x_i, y_i)$ corresponds to a pixel location in $I$ and $d_i\in\mathbb{R}$ is the propriocepted depth measurement at that location. For each $\mathbf{s}_i\in\mathrm{S}$, there exists an associated acquisition cost $c_i>0$. Let $D_\theta(I,\mathrm{S})$ be a Monocular Depth Estimator (MDE) that predicts a full depth map of the scene.

Given an uncertainty function $U(D_\theta(I,\mathrm{S}))$ that quantifies the predictive uncertainty of the MDE, our objective is to find S that minimizes the total acquisition cost while simultaneously minimizing uncertainty. This can be expressed as the minimization of the following constrained optimization problem:
\begin{equation}
\centering
    \min_{S\subset\Omega}\sum_{i=1}^Nc_i\quad\mathrm{s.t.}\quad U(D_\theta(I,\mathrm{S})) \leq \epsilon
\end{equation}
where $\epsilon$ is a target threshold for the uncertainty of the MDE and $\Omega$ represents the power set of all candidate pixel locations.


%% file: 4_Methodology.tex
\section{Methodology}

\subsection{Monocular Depth Estimation}

Our MDE is based on a U-Net architecture which takes as input the RGB image of a surgical scene as well as a sparse set of ground-truth depth measurements~\cite{ronneberger2015u}. The sparse measurements are represented as a depth map where unobserved pixels are set to -1, enabling the network to condition its predictions on available propriocepted depths. The MDE then predicts a full depth map of the scene, leveraging both visual features in the RGB image as well as the sparse depth information.

We train an ensemble of MDEs on simulation data where full ground-truth depth information is available. During training, sparse ground-truth sets are randomly sampled from the full depth map at varying densities, simulating the availability of partial propriocepted measurements. This encourages the network to incorporate known depth information when available while still learning to make accurate predictions from features in the RGB image. By training an ensemble with different initializations, we can measure the variance across predictions as a measure of model uncertainty which can subsequently be used to gather the most informative depth information.


\subsection{Uncertainty Quantification}

To quantify the uncertainty of the MDE ensemble, we first compute a pixelwise uncertainty map over the ensemble's set of predictions. Let $\{D_\theta^{(k)}(I,\mathrm{S})\}_{k=1}^K$ denote the outputs of $K$ independently trained MDE models, each conditioned on the RGB image $I$ and the sparse ground-truth set S. The predictive variance at each pixel $(x,y)$ is
\begin{equation}
    U_{x,y} = \frac{1}{K}\sum_{k=1}^K(D_\theta^{(k)}(I,\mathrm{S})_{x,y} - \bar{D}_{x,y})^2
\end{equation}
where $\bar{D}$ is the mean of the set of predictions,
\begin{equation}
    \bar{D} = \frac{1}{K}\sum_{k=1}^{K}D_\theta^{(k)}(I,\mathrm{S}).
\end{equation}

To obtain a scalar measure of overall uncertainty, we take the mean of the full variance map
\begin{equation}
U_{\mathrm{total}}=\frac{1}{N}\sum_{x,y}U_{x,y}
\end{equation}
where N is the total number of pixels in the image. This scalar represents the total predictive uncertainty of the ensemble over the entire predicted depth map.

Since $U_{\mathrm{total}}$ is differentiable with respect to the sparse depth measurements in S, we can compute $\nabla_\mathrm{S}U_{\mathrm{total}}$ which quantifies the expected reduction in ensemble uncertainty for each potential propriocepted depth measurement. We backpropagate through the network to the sparse depth map input, calculating the gradient of the variance average with respect to each pixel in the map. Locations corresponding to the largest-magnitude negative gradients indicate where acquiring depth information would most effectively reduce uncertainty. This gradient map therefore forms the foundation of our cost-aware active depth sensing framework, guiding the selection of depth measurements to maximize the improvement in predictive depth while minimizing the cost of gathering depth information.

\subsection{Active Depth Sensing}

After computing the gradient of the total ensemble uncertainty with respect to the sparse depth measurements, $\nabla_{\mathrm{S}}U_{\mathrm{total}}$, we use this information to actively select which locations to propriocept. Rather than selecting points greedily, we adopt Stein Variational Gradient Descent (SVGD) to optimize a set of candidate locations~\cite{liu2016stein}. Selecting points greedily could lead to a set of query points all near each other. SVGD strikes a balance between clustering points in highly informative regions while also encouraging diversity throughout the image. We first negate the gradient map so that the maximum values in the map correspond with locations with the highest likelihood of reducing uncertainty. The resulting map is then normalized to form a discrete probability distribution over all candidate pixel locations, indicating where measurements would be the most informative.

Formally, let $\{z_i\}_{i=1}^N$ denote a set of $N$ particles representing potential locations to propriocept. At each iteration, each particle is updated according to
\begin{equation}
    z_i \xleftarrow{} z_i + \eta\phi(z_i),
\end{equation}
where $\eta$ is the step size and $\phi(z_i)$ is the SVGD update function:
\begin{equation}
    \phi(z_i) = \frac{1}{N}\sum_{j=1}^N\bigg[k(z_j,z_i)(\nabla_{z_j}\log(p(z_j))+\nabla_{z_j}k(z_j,z_i)\bigg].
\end{equation}

Here, $p(z_i)$ is the discrete probability distribution obtained from the normalized negative gradient map, and $k(\cdot,\cdot)$ is the Radial Basis Function (RBF) kernel to encourage spatial diversity of the particles. This prevents all of the particles from collapsing to the same location, ensuring that the propriocepted locations are diverse throughout the scene.

Through these iterative updates, the particles converge to locations that maximize the expected reduction in total MDE uncertainty. The selected points are then propriocepted and added to the sparse ground-truth set S. A new prediction can then be made using this updated input to yield a full depth map that is more accurate in previously uncertain regions, while maintaining cost efficiency in the number of measurements acquired.

%% file: 5_Experiments.tex
\begin{figure}
    \centering
    \vspace{0.7em}
    \includegraphics[width=\columnwidth]{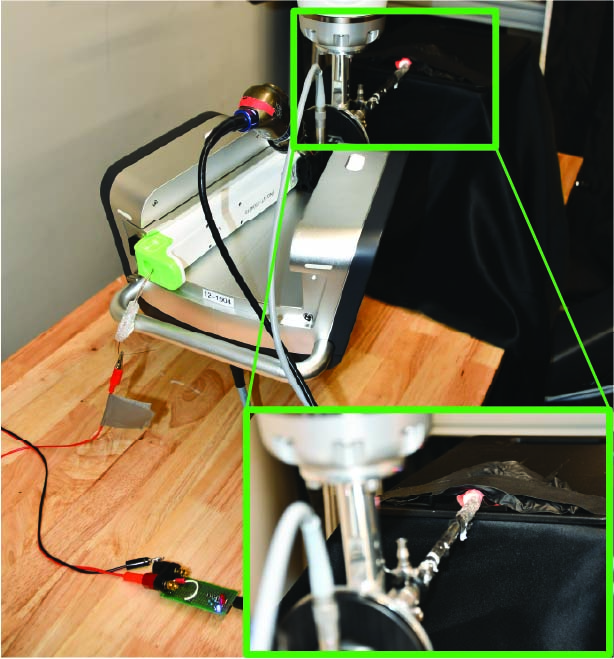}
    \caption{Physical experiment setup with Virtuoso Endoscopy System motor pack at center and capacitance sensing module at bottom left. A probe wire runs through the concentric tubes and extends from the tip of the robot. Detail view shows the endoscope sheath inserted into an ex vivo pig trachea.}
    \label{fig:physical_setup}
    \vspace{-1.5em}
\end{figure}

\section{Experiments}
\subsection{Datasets}

\textbf{Simulated Dataset:}
 A simulated dataset is generated from CT scans of 13 real-world ex vivo porcine tracheal models with an intraluminal tumor surrogate (chicken tissue) affixed using cyanoacrylate adhesive. We convert the scans into 3D meshes in Unity. A simulated camera explores the interior of the trachea reconstructions, collecting pairs of RGB images and depth maps. The RGB images are then translated into the style of real phantom endoscopic scenes using a style-transfer network \cite{park2020contrastive} to reduce the domain gap.

\textbf{Real-world Dataset:}
Our real-world dataset is collected from 12 other ex vivo porcine tracheal models with tumor surrogates prepared as above. We collect exploratory endoscopic videos of these models using the Virtuoso Endoscopy System (Virtuoso, Nashville, TN, USA) ~\cite{virtuoso-ves-manual} and then generate pseudo-ground-truth depth maps using Depth Anything V2's indoor large model \cite{yang_depth_2024} fine-tuned on the simulated dataset. We create additional data points by linearly and exponentially stretching the values of the pseudo-ground-truth depth maps. This artificially exaggerates the ambiguity of perceived scale, thereby forcing the model to rely heavily on the sparse ground truth. These pseudo-labels are used only to supervise training of the real-world model and are not used for evaluation.

\subsection{Ensemble Training}
We train an ensemble of $K = 3$ MDE models based on a U-Net backbone. Each model is trained independently with different random initializations on the same training and testing dataset. Training is performed using both the simulated and real-world datasets producing two ensembles (one for the simulated data experiments and one for physical trials). To train on sparse ground-truth sets, for each point in the dataset, we randomly sample 10 sparse ground-truth sets of varying densities with the following sampling strategies:
\begin{itemize}
    \item \textbf{Random Sampling:} Randomly sample a set number of points from the ground-truth depth map.
    \item \textbf{Cluster Based Sampling:} Randomly sample points in clusters from the ground-truth depth map.
    \item \textbf{Pixel Likelihood Sampling:} Each pixel location in the ground-truth depth map is included with a fixed probability.
\end{itemize}
These sampling strategies allow for the ensemble training to account for a wide range of potential sparse ground-truth sets. Details about the ensemble architecture can be seen in Fig.~\ref{fig:system_architecture}.

To allow for the ensemble's output to be unconstrained, each model outputs the log of the depth at each pixel location. Each model is then trained to minimize the scale-invariant loss between the predicted depth and the ground-truth depth:
\begin{equation}
\frac{1}{N}\sum_i(\log{d_i}-\log{\hat{d}_i)^2-\frac{1}{2N^2}\bigg(\sum_i(\log{d_i}-\log{\hat{d}_i})\bigg)^2}
\end{equation}
where $N$ is the total number of depth values and $d_i$, $\hat{d}_i$ are the ground-truth and predicted depth values respectively.

We train over $50$ epochs using an Adam optimizer with a learning rate of $3 \times 10^{-4}$. For each ensemble member we use the weights that demonstrate the lowest validation loss.

\subsection{Experimental Setup}
We first evaluate our approach on the simulated dataset. 
For each point in the simulated dataset, the MDE ensemble makes an initial prediction without access to any sparse ground-truth information. We then compare the performance of our method to several baseline point selection strategies.

To validate our framework on real-world data, we perform experiments using ex vivo porcine tracheal models with tumor surrogates using the Virtuoso Endoscopy System as seen in Fig. ~\ref{fig:physical_setup}. The system consists of a minimally invasive robot with concentric tube arms deployed through the working channel of an affixed rigid endoscope, pictured in Fig.~\ref{fig:front_page}. 


Whereas in the simulated setting, a dense ground-truth depth map was readily available, this information does not exist in the physical system. Consequently, depth estimates are obtained through a touch-sensing tool integrated at the distal end of the concentric tube arm. The touch-sensing tool, first introduced in \cite{Kumar_Hamlyn2025}, employs a capacitance circuit coupled to an electrosurgery probe running along the arm. Tissue contact produces a sharp change in capacitance, which can be measured in real time, enabling the arm to detect contact events. By using this tool to physically `probe' image-space points, we are able to recover their corresponding depths via robot kinematics.


During each trial, an RGB image of the surgical scene is acquired, and a set of desired measurement locations are selected using one of the point selection strategies (ours or a baseline for comparison). This supplies a list of pixel coordinates to probe for depth information. We use the calibrated camera intrinsics to back project each of the pixel coordinates as 3D rays whose origin is the camera position. The concentric tube arm, equipped with the touch-sensing tool, will iteratively trace along each of the rays until contact with tissue is sensed. The cartesian coordinate of this `sensed point' can then be used as the known depth information. 

Although the ray traced was calculated from the back projection of a specified pixel location, small errors may accrue from several sources. Thus, we re-project the sensed cartesian point back to a pixel coordinate on the image plane. The end result is a list of pixel coordinates with associated depth information. 
The MDE ensemble then receives the RGB image along with the set of selected depths and predicts a dense depth map of the scene.

\subsection{Baselines}
We compare our method against the following baseline point selection strategies:
\begin{itemize}
    \item \textbf{Random Selection:} Points are sampled uniformly at random across the image.
    \item \textbf{Greedy Variance:} The set of pixels with the highest ensemble predictive variance is selected.
    \item \textbf{Stein Variance:} SVGD particles are used over the ensemble variance map to select points.
    \item \textbf{Greedy Gradient:} The set of pixels with the largest magnitude in the gradient of the ensemble's uncertainty function is selected.
    \item \textbf{Ours (ProbeMDE):} Our proposed cost-aware SVGD method, where SVGD particles evolve over the normalized negated gradient of the ensemble variance.
\end{itemize}

Each method is evaluated with 5 points being added to the sparse ground-truth set at each iteration. For a fair comparison, each method is run over five iterations (total 25 points in the final prediction). As an additional baseline, we compare the reconstruction quality to a single MDE trained only on RGB image data (no sparse-ground truth).

\subsection{Evaluation Metrics}
To evaluate the reconstruction quality of each method, we report standard monocular depth estimation metrics widely used in prior work~\cite{eigen2014depth,eigen2015predicting,godard2017unsupervised,laina2016deeper}:
\begin{itemize}
    \item \textbf{Absolute Relative Error (AbsRel):} The mean absolute difference between predicted and ground-truth depth values, normalized by the ground truth: $\frac{1}{N}\sum_{i}\frac{|d_i-\hat{d}_i|}{d_i}$. This metric is sensitive to proportional errors across depth ranges. 
    \item \textbf{Squared Relative Error (SqRel):} The mean squared difference between predictions and ground truth, normalized by the ground truth: $\frac{1}{N}\sum_{i}\frac{(d_i-\hat{d}_i)^2}{d_i}$. This penalizes large outliers more heavily than AbsRel. 
    \item \textbf{RMSE:} The square root of the mean squared error between predicted and ground-truth depths: $\sqrt{\frac{1}{N}\sum_{i}(d_i-\hat{d}_i)^2}$. This metric emphasizes overall prediction accuracy. 
    \item \textbf{Log RMSE:} The RMSE computed in log space, which reduces the influence of absolute scale and highlights relative differences in depth predictions: $\sqrt{\frac{1}{N}\sum_{i}(\log{d_i}-log{\hat{d}_i})^2}$.
    \item \textbf{Scale-Invariant Error (ScInv):} A scale-invariant logarithmic loss that evaluates the consistency of depth relationships regardless of global scale shifts: \begin{equation*}\frac{1}{N}\sum_i(\log{d_i}-\log{\hat{d}_i)^2-\frac{1}{2N^2}\bigg(\sum_i(\log{d_i}-\log{\hat{d}_i})\bigg)^2}\end{equation*}.
    \item \textbf{Threshold Accuracy:} Fraction of pixels such that $\max(\frac{\hat{d}_i}{d_i},\frac{d_i}{\hat{d}_i})<1.25^t$ for $t\in(1,2,3)$. These thresholds evaluate prediction accuracy within progressively wider bounds.
\end{itemize}
Here, $d_i$ and $\hat{d}_i$ denote the ground-truth and predicted depth values, respectively, and $N$ is the total number of depth values in the depth map.

\begin{table*}[t]
\centering
\vspace{0.7em}
\begin{tabular}{lcccccccc}
\toprule
\textbf{Method} & \textbf{AbsRel} & \textbf{SqRel} & \textbf{RMSE} & \textbf{Log RMSE} & \textbf{ScInv} & $\delta < 1.25$ & $\delta < 1.25^2$ & $\delta < 1.25^3$ \\
\midrule
No Sparse Ground-Truth & 0.352 & 2.657 & 6.559 & 0.376 & 0.328 & 0.580 & 0.832 & 0.920 \\
Random Selection & 0.167 & 0.739 & 3.991 & 0.214 & 0.188 & 0.785 & 0.946 & 0.978 \\
Greedy Variance & 0.171 & 0.789 & 4.122 & 0.221 & 0.195 & 0.777 & 0.944 & 0.977 \\
Stein Variance & 0.170 & 0.789 & 4.182 & 0.221 & 0.195 & 0.776 & 0.942 & 0.976 \\
Greedy Gradient & 0.164 & 0.720 & \textbf{3.878} & 0.209 & 0.183 & 0.782 & 0.950 & \textbf{0.981} \\
ProbeMDE (Ours) & \textbf{0.162} & \textbf{0.708} & 3.920 & \textbf{0.206} & \textbf{0.180} & \textbf{0.791} & \textbf{0.952} & \textbf{0.981} \\
\bottomrule
\end{tabular}
\caption{Performance comparison of active proprioception strategies on simulated data. Results are reported across standard depth estimation metrics. Lower is better for AbsRel, SqRel, RMSE, Log RMSE, and ScInv. Higher is better for $\delta<1.25^t$. Best results in each column are bolded. We see that ProbeMDE performs consistently the strongest across the majority of metrics, highlighting our approach's ability to efficiently improve MDE performance.}
\label{tab:sim_results}
\end{table*}

\begin{table*}[t]
\centering
\begin{tabular}{lcccccccc}
\toprule
\textbf{Method} & \textbf{AbsRel} & \textbf{SqRel} & \textbf{RMSE} & \textbf{Log RMSE} & \textbf{ScInv} & $\delta < 1.25$ & $\delta < 1.25^2$ & $\delta < 1.25^3$ \\
\midrule
Random Selection & \textbf{0.242} & 3.745 & 14.254 & 0.391 & 0.331 & \textbf{0.452} & \textbf{0.809} & 0.939 \\
Greedy Variance & 0.261 & 3.695 & 13.797 & 0.385 & 0.316 & 0.377 & 0.774 & \textbf{0.944} \\
Stein Variance & 0.260 & 3.812 & 13.588 & 0.383 & 0.316 & 0.406 & 0.785 & 0.943 \\
Greedy Gradient & 0.264 & 4.021 & 14.503 & 0.410 & 0.341 & 0.392 & 0.767 & 0.936 \\
ProbeMDE (Ours) & 0.250 & \textbf{3.682} & \textbf{13.582} & \textbf{0.380} & \textbf{0.315} & 0.432 & 0.791 & 0.940 \\
\bottomrule
\end{tabular}
\caption{Performance comparison of active proprioception strategies on physical data. Results are reported across standard depth estimation metrics. Lower is better for AbsRel, SqRel, RMSE, Log RMSE, and ScInv. Higher is better for $\delta<1.25^t$. Best results in each column are bolded. These results illustrate how ProbeMDE performs consistently well across the majority of metrics, achieving the best performance in 4 metrics and second best in 3 other metrics.}
\label{tab:phys_results}
\vspace{-1.5em}
\end{table*}

%% file: 6_Results.tex
\begin{figure}
    \centering
    \includegraphics[width=\columnwidth]{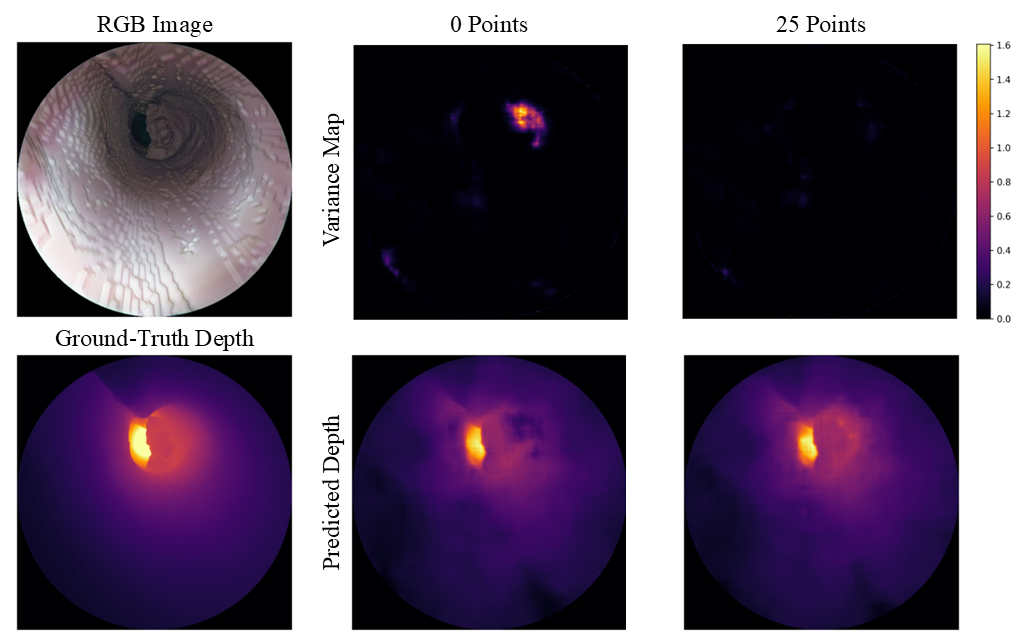}
    \caption{Qualitative results of ProbeMDE on a simulated central airway obstruction surgical scene. The left column shows the RGB image used as input to the ensemble as well as the ground-truth depth map that corresponds to the image. The middle and right columns show the ensemble's predictive variance along with the predicted depth map at 0 and 25 propriocepted depth measurements respectively. Qualitatively, the predicted depth maps get refined locally near propriocepted locations to improve detail in the depth map. We also see that as more points are propriocepted, there is a significant reduction in predictive variance across the depth map.}
    \label{fig:sim_qual}
    \vspace{-1.5em}
\end{figure}

\begin{figure}
    \centering
    \includegraphics[width=\columnwidth]{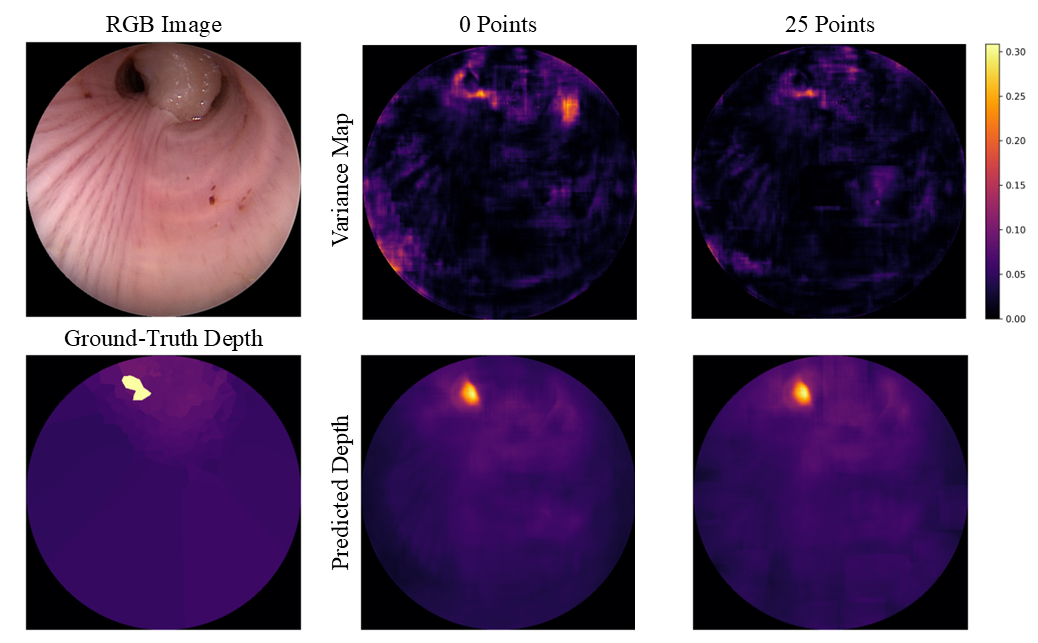}
    \caption{Qualitative results of ProbeMDE on a physical central airway obstruction surgical scene. The left column shows the RGB image used as input to the ensemble as well as the ground-truth depth map that corresponds to the image. As we do not have access to the full ground-truth depths in the case of our physical experiment, the ground-truth depth map shown is a Voronoi diagram constructed using known depth values at specific points within the image. The middle and right columns show the ensemble's predictive variance along with the predicted depth map at 0 and 25 propriocepted depth measurements respectively.}
    \label{fig:phys_qual}
    \vspace{-1.5em}
\end{figure}

\section{Results}
\subsection{Performance on Simulated Data}

Table~\ref{tab:sim_results} summarizes the results on our simulated dataset across all baselines and our method. Incorporating a small number of propriocepted ground-truth depths significantly improves depth predictions compared to solely relying on the RGB image. Specifically, the No Sparse Ground-Truth baseline exhibits substantially higher error across all metrics (e.g., SqRel = 2.657 and RMSE = 6.559) compared to any of the active-sensing strategies (e.g., ProbeMDE which achieved SqRel = 0.708 and RMSE = 3.920).

Among the active-sensing baselines, Random Selection already provides large gains, highlighting the value of even unstructured sparse propriocepted measurements. Greedy Variance and Stein Variance also provide significant gains over the no-depth method, but in this setting both are slightly worse than Random Selection. This suggests that targeting regions of high variance magnitude alone is not the most effective heuristic for guiding proprioceptive measurements.

The Greedy Gradient approach yields notable improvements, achieving the lowest RMSE (3.878) and tied for the best accuracy at $\delta < 1.25^3$ (0.981). However, our proposed ProbeMDE approach achieves the strongest overall performance, outperforming all baselines in all other metrics. This demonstrates that utilizing uncertainty gradients and optimizing proprioceptive measurement locations via SVGD provides a more principled and effective approach than variance-only or greedy selection methods.

Fig.~\ref{fig:sim_qual} shows a qualitative example of ProbeMDE on a simulated environment. In the case where there are no points in the sparse ground-truth set, we see regions of high variance in the variance map. As points are added to the sparse ground-truth set, we see a significant reduction in variance across the prediction while also improving the quality of the prediction, specifically in high depth regions.

Overall, these results confirm that actively guiding proprioception with our cost-aware Stein Gradient method leads to the largest performance improvements in monocular depth estimation accuracy over other active-sensing approaches.

\subsection{Physical Experiments}
We further evaluate our framework on physical data collected from surgical phantoms. In the absence of dense ground-truth for the physical case, we exhaustively probe the endoscopic scene to gather a sparse depth map on the order of $300$ points which we then use for evaluation. Each active sensing method selects a set of candidate query points which are projected to the nearest valid ground-truth location not already included in the input sparse ground-truth set, ensuring fairness across all selection strategies.

The quantitative results are summarized in Table~\ref{tab:phys_results}. Random Selection achieves the lowest AbsRel (0.242) and highest $\delta<1.25$ (0.452), indicating that selecting a set of uniform diverse points can be beneficial in this scenario. However, its overall performance is inconsistent across metrics with a relatively high RMSE (14.254), Log RMSE (0.391), and ScInv (0.331). Greedy and Stein Variance both improve $\delta <1.25^3$, but both methods perform poorly in most other metrics. The Greedy Gradient method performs the worst overall, showing relatively high errors across all metrics.

Our proposed ProbeMDE method achieves consistently strong performance across metrics, having the best SqRel (3.682), RMSE (13.582), Log RMSE (0.380), and ScInv (0.315) while also producing relatively low errors in the other metrics (second best in AbsRel, $\delta<1.25$, and $\delta<1.25^2$). The improvements that ProbeMDE provides across the majority of metrics highlight its robustness in the physical setting where ground-truth supervision is sparse.

Fig.~\ref{fig:phys_qual} shows an example of ProbeMDE on our physical experimental setup. Similarly to the simulated setting, we see a significant reduction in predictive variance as we increase the number of propriocepted points. Depth predictions also become locally more refined as we propriocept more points in the image.

%% file: 7_Conclusion.tex
\section{Conclusion}

We introduced ProbeMDE, a cost-aware active depth sensing framework that integrates sparse touch-based depth measurements with monocular depth estimation (MDE). By leveraging an ensemble of U-Net depth predictors, we quantify predictive uncertainty via ensemble variance and derive gradients that identify maximally informative locations for new proprioceptive measurements. Using Stein Variational Gradient Descent (SVGD), we select a diverse set of informative probing locations, enabling principled and efficient measurement acquisition.

Through both simulated and physical experiments, we demonstrate that ProbeMDE consistently improves depth estimation accuracy compared to baseline point selection strategies including random sampling, greedy heuristics, and gradient-based heuristics. Notably, our approach maximizes depth accuracy while minimizing the number of used proprioceptive measurements, highlighting the value of uncertainty-guided active sensing in cost-constrained robotic perception.

In future work, we plan to extend ProbeMDE to dynamic surgical environments, incorporating temporal consistency across video frames. Overall, our results illustrate that actively leveraging uncertainty to guide sparse measurements offers a scalable path toward safe and accurate perception in safety-critical robotics applications.